\documentclass{article}
\usepackage{graphicx}

\def\be{\begin{equation}}
\def\ee{\end{equation}}
\def\ff#1{\mbox{\boldmath $#1$} }

\def\lam{\lambda}
\def\x{\ff{x}}

\def\vcode#1#2#3#4{\begin{figure}
\begin{center}
\begin{minipage}[c]{#1\textwidth}
{{\small #2 \hrule \vspace{5pt}   %
{\it #3}  \vspace{5pt} \hrule }}
\end{minipage}
\caption{#4}
\end{center}   \end{figure}} 

\begin{document}

\title{Multi-objective Flower Algorithm for Optimization}

\author{Xin-She Yang, Mehmet Karamanoglu, \\
School of Science and Technology, \\
 Middlesex University, London NW4 4BT, UK.
 \and
Xingshi He \\
School of Science, Xi'an Polytechnic University, Xi'an, P. R. China.}

\date{}

\maketitle

\begin{abstract}
Flower pollination algorithm is a new nature-inspired algorithm, based on the characteristics
of flowering plants. In this paper, we extend this flower algorithm to solve multi-objective
optimization problems in engineering. By using the weighted sum method with random weights,
we show that the proposed multi-objective flower algorithm can accurately find the Pareto fronts
for a set of test functions. We then solve a bi-objective disc brake design problem,
which indeed converges quickly.
\end{abstract}

{\bf Citation detail:} X. S. Yang, M. Karamanoglu, X. S. He,
Multi-objective Flower Algorithm for Optimization, {\it Procedia Computer Science},
vol. 18, pp. 861-868 (2013).

\section{Introduction}

Engineering design optimization typically concerns multiple, often conflicting, objectives or multi-criteria,
and can thus be very challenging to solve. Therefore, some compromise and approximations are needed to
provide sufficiently good estimates to the true Pareto front of the problem of interest. Then, decision-makers
can rank different options, depending on their preferences or their utilities \cite{Abbass,Babu,Cag,Deb,Deb2000,Deb2,Reyes}.
In contrast with single objective optimization, multi-objective optimization has its additional
challenging issues such as time complexity, inhomogeneity and dimensionality.
To map the Pareto front accurately is very time-consuming, and there is no guarantee that these solutions
points will distribute uniformly on the front.
Single objective optimization typically has a single point in the solution space as its optimal solution,
while for bi-objective optimization; the Pareto front corresponds to a curve.  Higher dimensional problems can have
extremely complex hypersurface as its Pareto front \cite{Mada,Marler,Yang}. Consequently, these problems
can be extremely challenging to solve.

Nature-inspired algorithms have shown their promising performance and have thus become
popular and widely used, and these algorithms are mostly swarm intelligence based
\cite{Coello,Deb3,RAL,Yang,YangBA,YangBA11,YangBA12,Gandomi,GandomiYang}. These algorithms have also been
used to solve multiobjective optimization problems.

Therefore, the aim of this paper is to extend the flower pollination algorithm (FPA),
developed by Xin-She Yang in 2012 \cite{YangFPA}, for single objective optimization to
solve multiobjective optimization. The rest of this paper
is organized as follows: We first outline the basic characteristics of flower pollination and
then introduce in detail the ideas of flower pollination algorithm in Section 2. We then validate
the FPA by numerical experiments and a few selected multi-objective benchmarks in Section 3.
Then, in Section 4, we solve a real-world disc brake design benchmark with two objectives.
Finally, we discuss some relevant issues and conclude in Section 5.

\section{Nature-Inspired Flower Pollination Algorithm}

\subsection{Pollination of Flowering Plants}

Flowering plant has been evolving for at least more than 125 million years.
 It is estimated that there are over a quarter of a million types of flowering plants in Nature
and that about 80\% of all plant species are flowering species. It still remains
a mystery how flowering plants came to dominate the landscape from
Cretaceous period \cite{Walker}.
The primary purpose of a flower is ultimately reproduction via pollination.
Flower pollination is typically associated with the transfer of pollen, and such transfer
is often linked with pollinators such as insects, birds, bats and other animals.
In fact, some insects and certain flowers have co-evolved into a very specialized
flower-pollinator partnership. For example, some flowers can only depend on
a specific species of insects or birds for successful pollination.

Abiotic and biotic pollination are two main forms in the pollination process.  About 90\%
of flowering plants belong to biotic pollination. That is, pollen is
transferred by a pollinator such as insects and animals. About 10\%
of pollination takes abiotic form which does not require any pollinators.
Wind and diffusion help pollination of such flowering plants,
and grass is a good example of abiotic pollination \cite{Flower,Glover}.
Pollinators, or sometimes called pollen vectors, can be very diverse.
It is estimated there are at least about 200,000 varieties of pollinators such as
insects, bats and birds. Honeybees are a good example of pollinator, and they
have also developed the so-called flower constancy. That is,
these pollinators tend to visit exclusive certain flower species while bypassing
other flower species. Such flower constancy may have evolutionary advantages
because this will maximize the transfer of flower pollen to the same or conspecific plants,
and thus maximizing the reproduction of the same flower species.
Such flower constancy may be advantageous for pollinators as well, because they
can be sure that nectar supply is available with their limited memory and minimum
cost of learning, switching or exploring. Rather than focusing on some unpredictable but potentially
more rewarding new flower species, flower constancy may require minimum investment
cost and more likely guaranteed intake of nectar \cite{Waser}.

By a close look into the world of flowering plants,
pollination can be achieved by self-pollination or cross-pollination.
Cross-pollination, or allogamy, means pollination can occur from pollen
of a flower of a different plant, while self-pollination is the fertilization
of one flower, such as peach flowers, from pollen of the same flower or different flowers of the same plant,
which often occurs when there is no reliable pollinator available.
Biotic, cross-pollination may occur at long distance, and the pollinators such as
bees, bats, birds and flies can fly a long distance, thus they can considered as the
global pollination. In addition, bees and birds may behave as L\'evy flight behaviour \cite{Pav},
with jump or fly distance steps obeying a L\'evy distribution. Furthermore,
flower constancy can be considered as an increment step using the similarity or difference
of two flowers. From the biological evolution point of view, the objective of the flower pollination
is the survival of the fittest and the optimal reproduction of plants in terms of
numbers as well as the most fittest.

\subsection{Flower Pollination Algorithm}

Based on the above characteristics of flower pollination,
Xin-She Yang developed the Flower pollination algorithm (FPA) in 2012 \cite{YangFPA}.
For simplicity, we use the following four rules:

\begin{enumerate}

\item Biotic and cross-pollination can be considered as a process of
global pollination process, and pollen-carrying pollinators move in a way which
obeys L\'evy flights (Rule 1).

\item For local pollination, abiotic and self-pollination are used (Rule 2).

\item Pollinators such as insects can develop flower constancy, which is equivalent to
a reproduction probability that is proportional to the similarity of two flowers involved (Rule 3).

\item The interaction or switching of local pollination and global pollination can
be controlled by a switch probability $p \in [0,1]$, with a slight bias towards local
pollination (Rule 4).

\end{enumerate}

From the implementation point of view, a set of updating formulae are needed.
Now we convert the above rules into updating equations. First, in the global pollination step,
flower pollen gametes are carried by pollinators such as insects,
and pollen can travel over a long distance because insects can often fly and move in a much
longer range. Therefore, Rule 1 and flower constancy can be  represented mathematically as
\be \x_i^{t+1}=\x_i^t + \gamma L(\lam) (\x_i^t -\ff{g}_*), \ee
where $\x_i^t$ is the pollen $i$ or solution vector $\x_i$ at iteration $t$, and $\ff{g}_*$ is
the current best solution found among all solutions at the current generation/iteration.
Here $\gamma$ is a scaling factor to control the step size.
In addition, $L(\lam)$ is the parameter that corresponds to the strength of the pollination,
which essentially is also the step size.
Since insects may move over a long distance with various distance steps, we can use a L\'evy flight
to mimic this characteristic efficiently. That is, we draw $L>0$ from a Levy distribution
 \be L \sim \frac{\lam \Gamma(\lam) \sin (\pi \lam/2)}{\pi} \frac{1}{s^{1+\lam}}, \quad (s \gg s_0>0). \ee
Here, $\Gamma(\lam$) is the standard gamma function, and this distribution is valid for
large steps $s>0$.

\vcode{0.9}{{\sf Flower Pollination Algorithm (or simply Flower Algorithm) }} {
\indent Objective $\min$ or $\max f(\x)$, $\x=(x_1,x_2,..., x_d)$ \\
\indent Initialize a population of $n$ flowers/pollen gametes with random solutions  \\
\indent Find the best solution $\ff{g}_*$ in the initial population \\
\indent Define a switch probability $p \in [0,1]$ \\
\indent Define a stopping criterion (either a fixed number of generations/iterations or accuracy) \\
\indent  {\bf while} ($t<$MaxGeneration) \\
\indent \qquad {\bf for} $i=1:n$ (all $n$ flowers in the population) \\
\indent \quad \qquad {\bf if} rand $<p$, \\
\indent \qquad \qquad Draw a ($d$-dimensional) step vector $L$ which obeys a L\'evy distribution \\
\indent \qquad \qquad Global pollination via $\x_i^{t+1}=\x_i^t + L (\ff{g}_*-\x_i^t)$ \\
\indent \qquad \quad {\bf else} \\
\indent \qquad \qquad Draw $\epsilon$ from a uniform distribution in [0,1] \\
\indent \qquad  \qquad Do local pollination via $\x_i^{t+1}=\x_i^t+\epsilon (\x_j^t-\x_k^t)$ \\
\indent \qquad \quad {\bf end if } \\
\indent \qquad \quad Evaluate new solutions \\
\indent \qquad \quad If new solutions are better, update them in the population \\
\indent \qquad {\bf end for} \\
\indent \qquad \quad Find the current best solution $\ff{g}_*$ \\
\indent  {\bf end while}  \\
\indent Output the best solution found }
{Pseudo code of the proposed Flower Pollination Algorithm (FPA). \label{fpa-code} }

Then, to model the local pollination, both Rule 2 and Rule 3 can be represented as
\be \x_i^{t+1}=\x_i^t + \epsilon (\x_j^t -\x_k^t), \ee
where $\x_j^t$ and $\x_k^t$ are pollen from different flowers of the same plant species.
This essentially mimics the flower constancy in a limited neighbourhood. Mathematically,
if $\x_j^t$ and $\x_k^t$ comes from the same species
or selected from the same population, this equivalently becomes a local random walk if we draw
$\epsilon$ from a uniform distribution in [0,1].

Though Flower pollination activities can occur at all scales, both local and global,
adjacent flower patches or flowers in the not-so-far-away neighbourhood
are more likely to be pollinated by local flower pollen than those far away.
In order to mimic this, we can effectively use a switch probability (Rule 4) or proximity probability $p$ to switch
between common global pollination to intensive local pollination.
To start with, we can use a naive value of $p=0.5$ as an initially value.
A preliminary parametric showed that $p=0.8$ might work better for most applications.

\subsection{Multi-objective Flower Pollination Algorithm (MOFPA)}

There are quite a few approaches to dealing multi-objectives using algorithms that have been tested
by single-objective optimization problems. Perhaps, the simplest way is to use
a weighted sum to combine all multiple objectives into a composite single objective
\be f=\sum_{i=1}^m w_i f_i, \quad \sum_{i=1}^m w_i=1, \quad w_i>0, \ee
where $m$ is the number of objectives and $w_i (i=1,...,m)$ are non-negative weights.
In order to obtain the Pareto front accurately with solutions uniformly distributed
on the front, we have to use random weights $w_i$, which can be drawn
from a uniform distribution, or low-discrepancy random numbers.

\section{Simulation and Results}

Various test functions for multi-objective optimization exist \cite{ZhangZhou,Zit,Zitz},
though there is no agreed set available at present. Ideally, a new algorithm should be
tested again all known test functions, however, this is a time-consuming task.
In practice, we often use a subset of some widely used functions with diverse
properties of Pareto fronts. To validate the proposed MOFA, we have selected a subset
of these functions with convex, non-convex and discontinuous Pareto fronts.
We will first use four test functions, and then solve a bi-objective disc brake
design problem.

The parameters in MOFPA are fixed in the rest of the paper, based on
a preliminary parametric study, and we will use $p=0.8$, $\lam=1.5$ and
a scaling factor $\gamma=0.1$. The population size $n=50$ and the number of
iterations is set to $t=1000$.

\subsection{Test Functions}

For simplicity in this paper, we have tested the following four functions:
\begin{itemize}
\item ZDT1 function with a convex front \cite{Zit,Zitz}
\[ f_1(x)=x_1, \quad f_2(x)=g (1-\sqrt{f_1/g}), \]
\be g=1+\frac{9 \sum_{i=2}^d x_i}{d-1}, \quad x_1 \in [0,1], \; i=2,...,30, \ee
where $d$ is the number of dimensions. The Pareto-optimality is reached when $g=1$.

\item ZDT2 function with a non-convex front
\[ f_1(x)=x_1, \quad f_2(x) =g (1-\frac{f_1}{g})^2, \]

\item ZDT3 function with a discontinuous front
\[ f_1(x) =x_1, \quad f_2(x)=g \Big[1-\sqrt{\frac{f_1}{g}}-\frac{f_1}{g} \sin (10 \pi f_1) \Big], \]
where $g$ in functions ZDT2 and ZDT3 is the same as in function ZDT1. In the ZDT3 function,
$f_1$ varies from $0$ to $0.852$ and $f_2$ from $-0.773$ to $1$.

\item LZ function \cite{Li,ZhangQ}
\[ f_1=x_1 +\frac{2}{|J_1|} \sum_{j \in J_1} \Big [ x_j -\sin (6 \pi x_1 +\frac{j \pi}{d}) \Big]^2, \]
\be f_2=1-\sqrt{x_1} + +\frac{2}{|J_2|} \sum_{j \in J_2} \Big [ x_j -\sin (6 \pi x_1 +\frac{j \pi}{d}) \Big]^2, \ee
where $J_1=\{j|j$ is odd $\}$ and $J_2 =\{ j|j$ is even $\}$ where $2 \le j \le d$. This
function has a Pareto front $f_2=1-\sqrt{f_1}$ with a Pareto set
\be x_j=\sin (6 \pi x_1 + \frac{j \pi}{d}), \quad j=2,3, ..., d, \quad x_1 \in [0,1]. \ee

\end{itemize}

\begin{figure}
\centerline{\includegraphics[height=4in,width=5in]{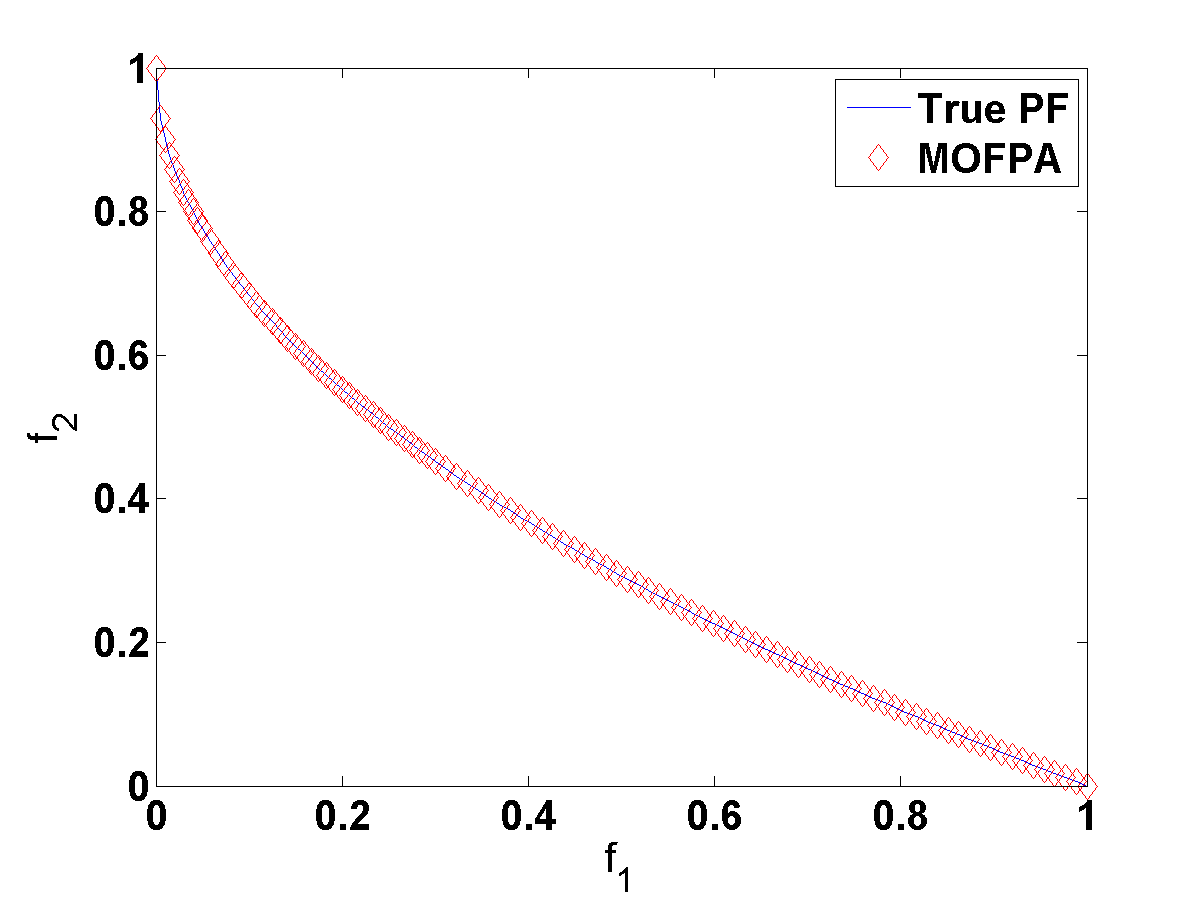}}
\caption{Pareto front of test function ZDT1. \label{fig-300}}
\end{figure}

We first generated 100 Pareto points by MOFPA, and then compared the Pareto front generated by MOFPA
with the true front $f_2=1-\sqrt{f_1}$ of ZDT1, and the results are shown in Fig. \ref{fig-300}.

Let us define the distance or error between the estimated Pareto front $PF^e$ to
its corresponding true front $PF^t$ as
\be E_f=||PF^e-PF^t||^2=\sum_{j=1}^N (PF_j^e-PF_j^t)^2, \ee
where $N$ is the number of points. The convergence property can be viewed by
following the iterations. As this measure is an absolute measure, which depends on
the number of points. Sometimes, it is easier to use relative measure using generalized
distance
\be D_g=\frac{1}{N} \sqrt{\sum_{j=1}^N (PF_j-PF_j^t)^2}. \ee

\subsection{Performance Comparison}

To see how the proposed algorithm performs in comparison with other algorithms, we
now compare the performance of the proposed MOFPA with other established
multiobjective algorithms. Not all algorithms have extensive published results,
so we have carefully selected a few algorithms with available results from the literature.
In case of the results are not available,
we have tried to implement the algorithms using well-documented studies
and then generated new results using these algorithms. In particular, we have used other
methods for comparison, including vector evaluated genetic algorithm
(VEGA) \cite{Schaf}, NSGA-II \cite{Deb3}, multi-objective differential evolution (MODE) \cite{Xue,Babu},
differential evolution for multi-objective optimization (DEMO) \cite{Robic},
multi-objective bees algorithms (Bees) \cite{Pham},
and Strength Pareto Evolutionary Algorithm (SPEA) \cite{Deb3,Mada}. The performance measures in terms of
generalized distance $D_g$ are summarized in Table 1 for all
the above major methods.

From Table 1, we can see that the proposed MOFPA obtained better
results for almost all four cases.

\begin{table}[h]
\caption{Comparison of $D_g$ for $n=50$ and $t=500$ iterations. }
\begin{center} \begin{tabular}{|l|l|l|l|l|l|l|}
\hline
Methods &  ZDT1 & ZDT2 & ZDT3 & LZ \\
\hline \hline
VEGA & 3.79E-02 & 2.37E-03  & 3.29E-01 &   1.47E-03  \\
NSGA-II & 3.33E-02 & 7.24E-02  & 1.14E-01    & 2.77E-02  \\
MODE & 5.80E-03 & 5.50E-03 &     2.15E-02 & 3.19E-03  \\
DEMO & 1.08E-03 & 7.55E-04 &  1.18E-03 &    1.40E-03 \\
Bees  & 2.40E-02 & 1.69E-02 & 1.91E-01     & 1.88E-02 \\
SPEA & 1.78E-03 & 1.34E-03 & 4.75E-02   & 1.92E-03  \\ \hline
MOFPA & 7.11E-05 & 1.24E-05 & 5.49E-04   & 7.92E-05 \\
\hline
\end{tabular} \end{center} \end{table}

\section{Design of a Disc Brake With Two Objectives}
There are a few dozen benchmarks in the engineering literature \cite{Kim,Pham,RAL,Rang}.
We now use the MOFPA to solve a disc brake design benchmark \cite{Gong,Osy,RAL}.
The objectives are to minimize the overall mass and the braking time by choosing optimal design
variables: the inner radius $r$, outer radius $R$ of the discs, the engaging force $F$ and the number
of the friction surface $s$. This is under the design constraints such as the torque, pressure, temperature,
and length of the brake. This bi-objective design problem can be written as:
\be \textrm{Minimize } \; f_1(\x) =4.9 \times 10^{-5} (R^2-r^2) (s-1),
\quad f_2(\x)=\frac{9.82 \times 10^6 (R^2-r^2)}{F s (R^3-r^3)}, \ee
subject to
\be
\begin{array}{lll}
g_1(\x) =  20-(R-r) \le 0, \\[15pt]
g_2(\x) = 2.5 (s+1)-30 \le 0, \\[15pt]
g_3(\x) = \frac{F}{3.14 (R^2-r^2)} -0.4 \le 0, \\[15pt]
g_4(\x) = \frac{2.22 \times 10^{-3} F (R^3 -r^3)}{(R^2-r^2)^2} -1 \le 0, \\[15pt]
g_5(\x) =900- \frac{0.0266 F s (R^3-r^3)}{(R^2-r^2)} \le 0.
\end{array}
\ee
The simple limits are \be 55 \le r \le 80, \; 75 \le R \le 110, \; 1000 \le F \le 3000, \; 2 \le s \le 20. \ee
It is worth pointing out that $s$ is discrete. In general, we have to extend MOFPA in combination with
constraint handling techniques to
deal with mixed integer problems efficiently. However, since there is only one discrete variable,
we can use the simplest branch-and-bound method.

\begin{figure}[h]
\centerline{\includegraphics[height=4.25in,width=5.5in]{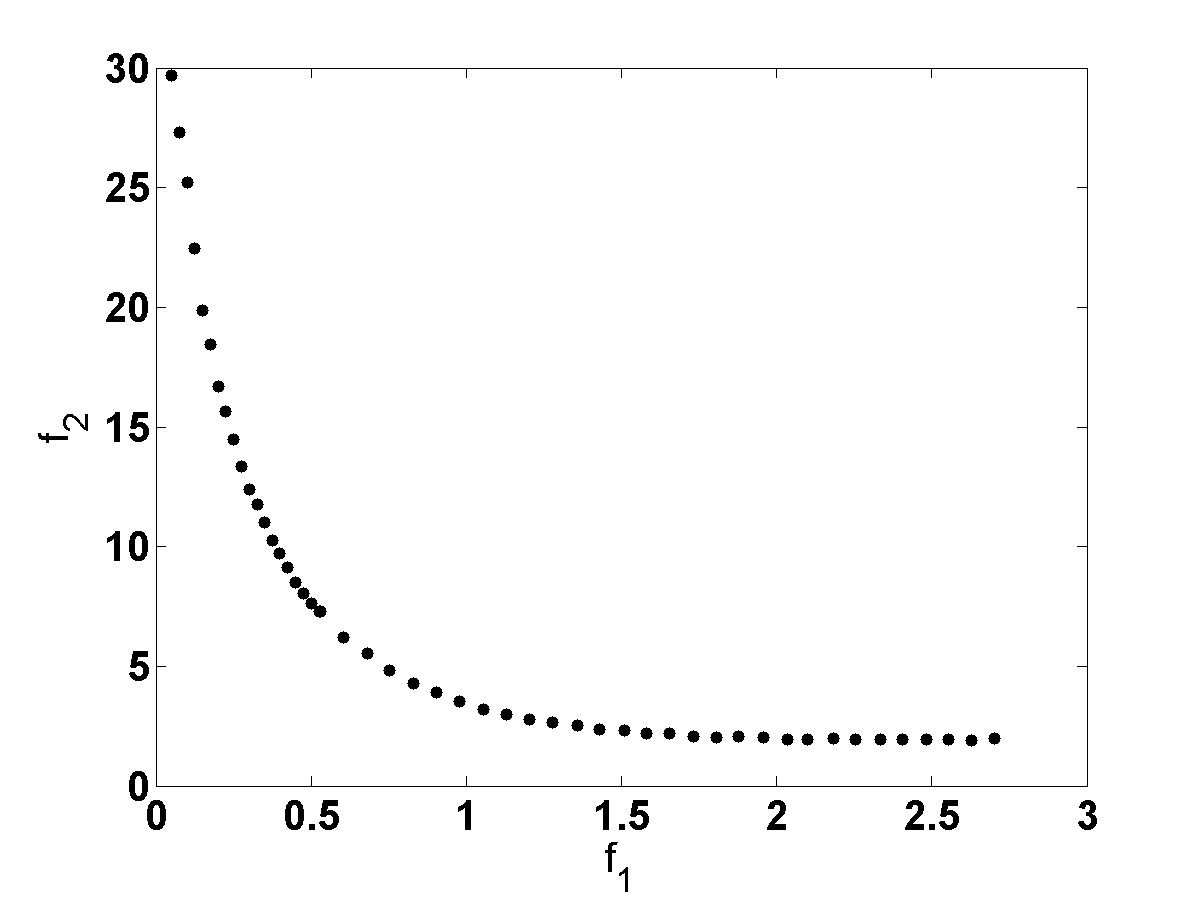} }
\caption{Pareto front of the disc brake design. \label{fig-800}}
\end{figure}

The above results for these benchmarks and test functions suggest that MOFPA is a very efficient
algorithm for multi-objective optimization. It can deal with highly nonlinear problems
with complex constraints and diverse Pareto optimal sets.

\section{Conclusions}
We have successfully extended a flower algorithm for single-objective optimization to
solve multi-objective design problems. Numerical experiments and design benchmarks
have shown that MOFPA is very efficient with an almost exponential convergence rate.
This observation is based on the comparison of FPA with other algorithms for
solving multi-objective optimization problems.

The standard FPA has its simplicity and flexibility, and in many ways, it has
some similarity to that of cuckoo search and other algorithms with L\'evy flights \cite{Yang,YangRA}.
FPA has only one key parameter $p$ together with a scaling factor $\gamma$,
which makes the algorithm easier to implement.

It is worth pointing out that we have only done some preliminary parametric studies.
Future studies can focus on more detailed parametric analysis and their possible links with performance.
Furthermore, the linearity in the main updating formulas makes it possible to do some theoretical
analysis in terms of dynamic systems or Markov chain theories. This could form an useful
topic for further research.

\end{document}